\documentclass[conference]{IEEEtran}
\IEEEoverridecommandlockouts
\usepackage{cite}
\usepackage{amsmath,amssymb,amsfonts}
\usepackage{algorithmic}
\usepackage{graphicx}
\usepackage{hyperref}
\usepackage{textcomp}
\usepackage[table]{xcolor}
\usepackage{booktabs}
\usepackage{amsmath} 
\usepackage{multicol}
\usepackage{multirow}
\def\BibTeX{{\rm B\kern-.05em{\sc i\kern-.025em b}\kern-.08em
    T\kern-.1667em\lower.7ex\hbox{E}\kern-.125emX}}
\begin{document}

\title{Dual Semantic-Aware Network for Noise Suppressed Ultrasound Video Segmentation\\
}

\author{
Ling Zhou, Runtian Yuan, Yi Liu, Yuejie Zhang, Rui Feng, Shang Gao
\thanks{This work is supported by Shanghai Natural Science Foundation (No. 25ZR1401028), and the Science and Technology Commission of Shanghai Municipality (No. 23511100602; No. 21511104506).}
\thanks{Ling Zhou, Runtian Yuan, Yuejie Zhang and Rui Feng are with the College of Computer Science and Artificial Intelligence Shanghai Key Laboratory of Intelligent Information Processing, Fudan University, Shanghai, China (e-mail: lzhou24@m.fudan.edu.cn; rtyuan21@m.fudan.edu.cn; yjzhang@fudan.edu.cn; fengrui@fudan.edu.cn)}
\thanks{Yi Liu is with the School of Computer Science and Artificial Intelligence, the Aliyun School of Big Data, the School of Software, and the CNPC-CZU Innovation Alliance, Changzhou University, Changzhou, Jiangsu, China (e-mail: liuyi0089@gmail.com)}
\thanks{Shang Gao is with the School of Information Technology, Deakin University, Waurn ponds, Australia (email:shang@deakin.edu.au)}
}

\maketitle

\begin{abstract}
Ultrasound imaging is a prevalent diagnostic tool known for its simplicity and non-invasiveness. However, its inherent characteristics often introduce substantial noise, posing considerable challenges for automated lesion or organ segmentation in ultrasound video sequences. To address these limitations, we propose the Dual Semantic-Aware Network (DSANet), a novel framework designed to enhance noise robustness in ultrasound video segmentation by fostering mutual semantic awareness between local and global features. Specifically, we introduce an Adjacent-Frame Semantic-Aware (AFSA) module, which constructs a channel-wise similarity matrix to guide feature fusion across adjacent frames, effectively mitigating the impact of random noise without relying on pixel-level relationships. Additionally, we propose a Local-and-Global Semantic-Aware (LGSA) module that reorganizes and fuses temporal unconditional local features, which capture spatial details independently at each frame, with conditional global features that incorporate temporal context from adjacent frames. This integration facilitates multi-level semantic representation, significantly improving the model's resilience to noise interference. Extensive evaluations on four benchmark datasets demonstrate that DSANet substantially outperforms state-of-the-art methods in segmentation accuracy. Moreover, since our model avoids pixel-level feature dependencies, it achieves significantly higher inference FPS than video-based methods, and even surpasses some image-based models. Code can be found in \href{https://github.com/ZhouL2001/DSANet}{DSANet}

\end{abstract}


\section{Introduction}

\begin{figure}[t]
\includegraphics[width=\linewidth]{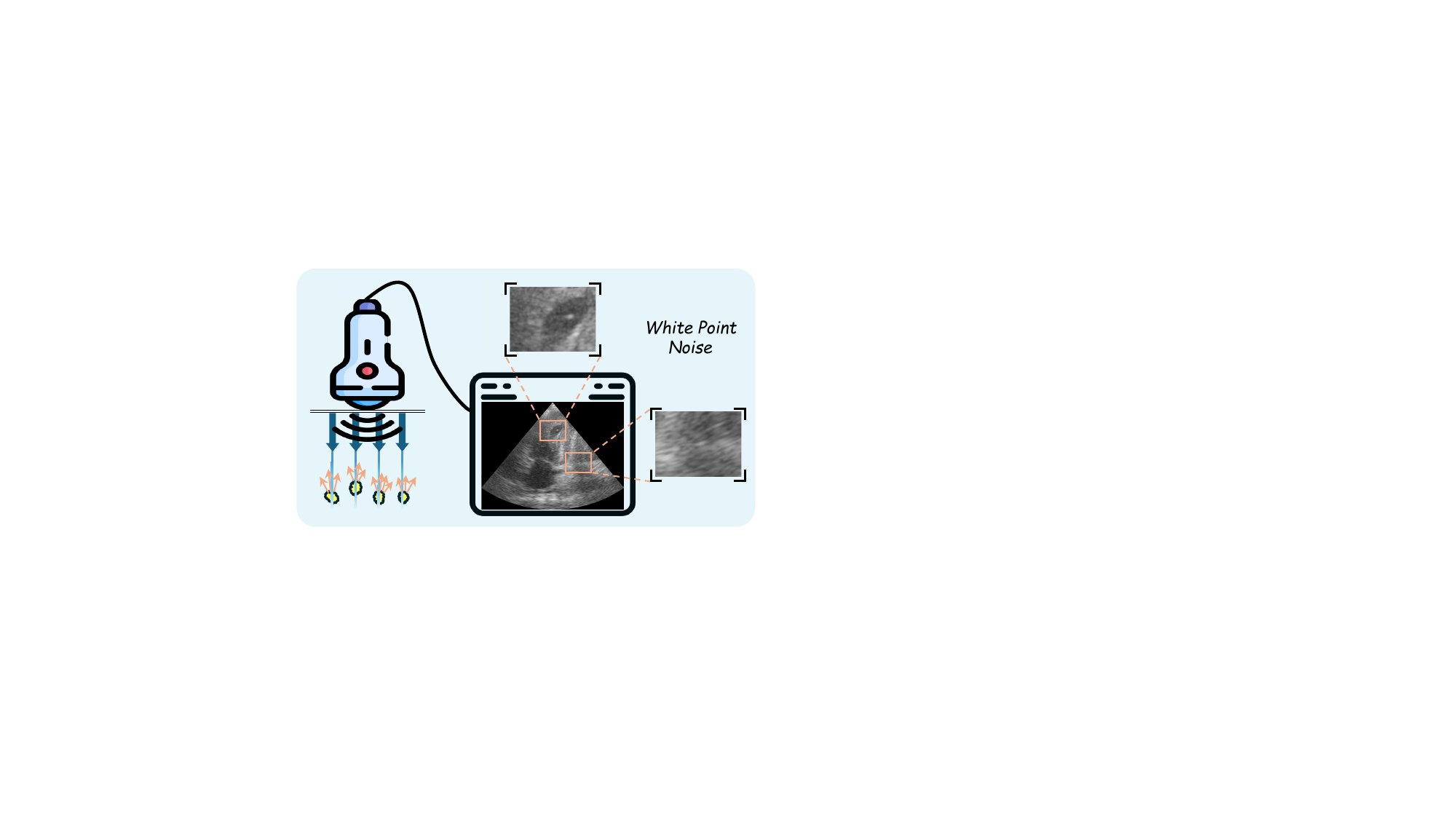}
\caption{Ultrasound images are formed from the echoes of coherent acoustic waves. When these waves interact with sub‑wavelength scatterers inside tissue, they create myriad microscopic scattering centres whose constructive and destructive interference manifests as granular, point‑like speckle noise.} \label{ultrasound}
\end{figure}

Ultrasound imaging is widely used in clinical diagnosis due to its simplicity and non-invasive nature. A key task in this field is the detection of organs and lesions, which is essential for accurate diagnosis and treatment planning. However, ultrasound imaging relies on the transmission of coherent acoustic pulses and the reception of their back-scattered echoes \cite{zemp2005detection}. As these pulses traverse biological tissues containing innumerable sub-wavelength scatterers, constructive and destructive interference among echoes gives rise to granular, point-like speckle noise \cite{shin2016spatial}, as illustrated in Fig. \ref{ultrasound}. These noise not only reduces image contrast and obscures fine anatomical detail, but poses significant challenges to robust, automated spatio-temporal segmentation in dynamic ultrasound sequences.

Recent advances in deep learning, ranging from convolutional neural networks to attention-based transformers \cite{vaswani2017attention}, have been increasingly utilized for ultrasound video segmentation \cite{jiang2024towards, lin2023shifting, li2022rethinking}. These methods generally follow a two-stage pipeline: the first stage establishes temporal relationships between adjacent frames to generate temporally conditioned features, while the second stage uses these conditional features to produce final segmentation predictions. Although these methods have achieved promising results, they often fail to address the diverse effects of inherent noise in ultrasound images during both stages. 

\begin{figure}[t]
\includegraphics[width=\linewidth]{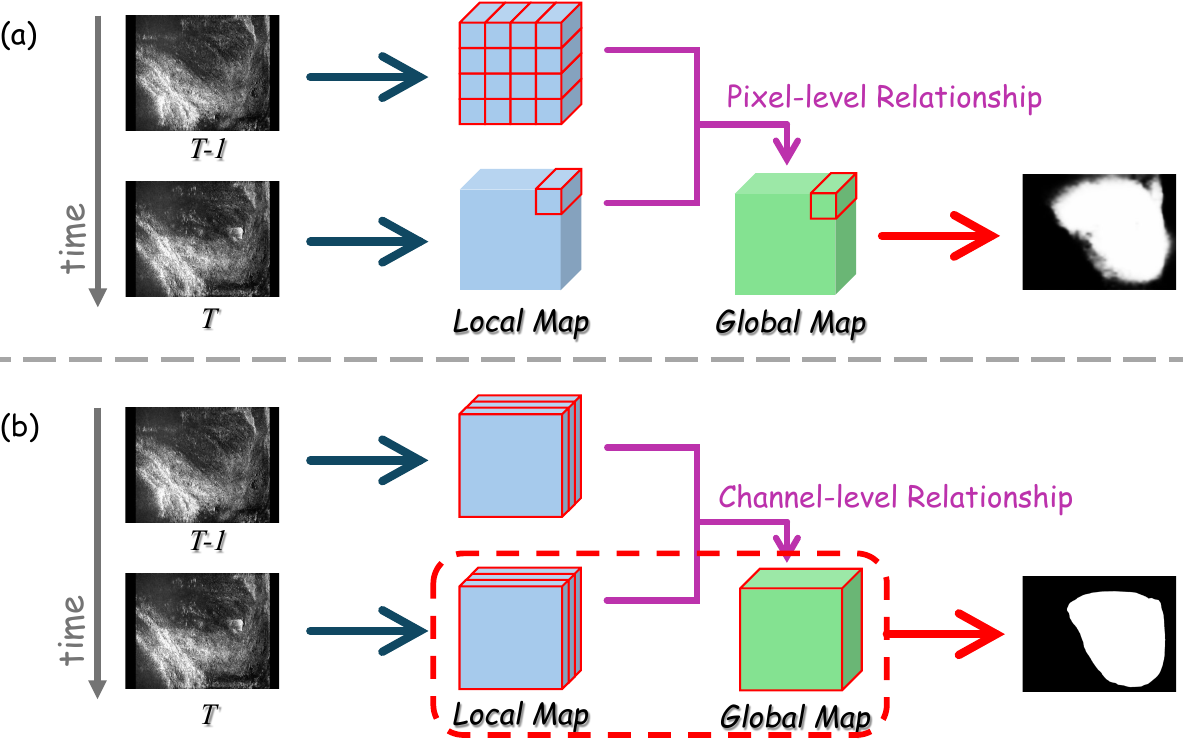}
\caption{Comparison between prior work and our method. (a) Previous methods build inter-frame relationships at the pixel level, enhancing the local feature map into a temporally conditioned global map, which is then used to generate the final prediction. (b) Our method builds inter-frame relationships at the channel level and jointly utilizes both local and global maps to produce the final segmentation.} \label{statement}
\end{figure}

Specifically, in the first stage, existing methods \cite{gong2023thyroid, fan2020pranet} commonly rely on pixel-level computations, such as cross-attention mechanism, to model relationships between adjacent frames. As shown in Fig. \ref{statement} (a), these methods match individual pixels across frames, which can indiscriminately treat noisy pixels as valid. 
This not only amplifies the impact of random noise, degrading feature discriminability and temporal coherence, but also leads to inaccurate and unstable segmentation predictions. Furthermore, the pixel-level complexity imposes a heavy computational burden, hindering deployment in clinical diagnostic settings. 

In the second stage (Fig. \ref{statement} (a)), most approaches\cite{jiang2024towards, li2022rethinking} fuse features from initial frames with the local features of the target frame to produce temporally conditioned global features \footnote{
Note: In this paper, local features refer to features of the target frame without temporal fusion, while global features include temporal information from adjacent frames.}. 
These global features are then used to make the final segmentation. 
While these methods utilize temporal context to locate dynamic targets, they often overlook the holistic information embedded in local features, which is inevitably weakened during temporal integration. Prior work \cite{zheng2020cross, liu2023tcgnet} has demonstrated that preserving this holistic information can reduce noise interference and enhance the robustness.

To address these challenges, we propose a novel two-stage framework named \textbf{D}ual \textbf{S}emantic-\textbf{A}ware Network (DSANet). The term `dual' emphasizes two types of semantic interactions: (1) between adjacent-frame semantics for temporal consistency, and (2) between local and global semantics for multi-scale feature representation. 

In the first stage, we introduce the \textbf{A}djacent-\textbf{F}rame \textbf{S}emantic-\textbf{A}ware (AFSA) module. As shown in Fig. \ref{statement} (b), this module leverages cosine similarity to capture semantic relationships between adjacent frames at the channel level. By guiding feature refinement using inter-frame channel similarities, AFSA improves semantic consistency while suppressing noise. 
Two pooling operations are then applied to derive fine-grained feature maps. Notably, all operations are conducted at the channel level, which enhances feature stability by preserving spatial consistency and mitigating the influence of noise and local intensity fluctuations. 

In the second stage, we introduce the \textbf{L}ocal-and-\textbf{G}lobal \textbf{S}emantic-\textbf{A}ware (LGSA) module. 
Instead of relying solely on unconditional local features for saliency prediction, the LGSA module simultaneously incorporates unconditional local features to complement holistic representation information. Inspired by \cite{prajapati2021channel, liu2023tcgnet}, which have proven that channel-level fusion exhibits robust anti-interference performance, 
we design a channel splitting and reassembly strategy. By separating and interactively recombining channels from local and global features, we enable cross-scale interactions that allow fine-grained details to benefit from global context while preserving crucial spatial information. Given the varied granularity of noise patterns, this 
structured fusion strategy effectively suppresses irrelevant noise while preserving crucial feature details, leading to more robust and discriminative representations for final prediction.

In summary, our main contributions are as follows:
\begin{enumerate}
    \item We introduce a novel framework named DSANet, for noise-robust ultrasound video lesion segmentation, which promotes semantic interaction at the channel level to significantly reduce the impact of intrinsic noise.
    \item We develop an Adjacent-Frame Semantic-Aware module to capture semantic similarity across frames using channel-level interactions, reducing noise impact and computational cost. 
    \item We design a Local-and-Global Semantic-Aware module, which integrates unconditional local and conditional global features though channel splitting and reassembly, improving noise robustness and prediction quality.
    \item Extensive experiments on four benchmark datasets demonstrate that DSANet outperforms previous state-of-the-art methods, and the inference FPS is higher than video-based methods, even surpassing some image-based methods.
\end{enumerate}

\begin{figure*}[t]
\includegraphics[width=\textwidth]{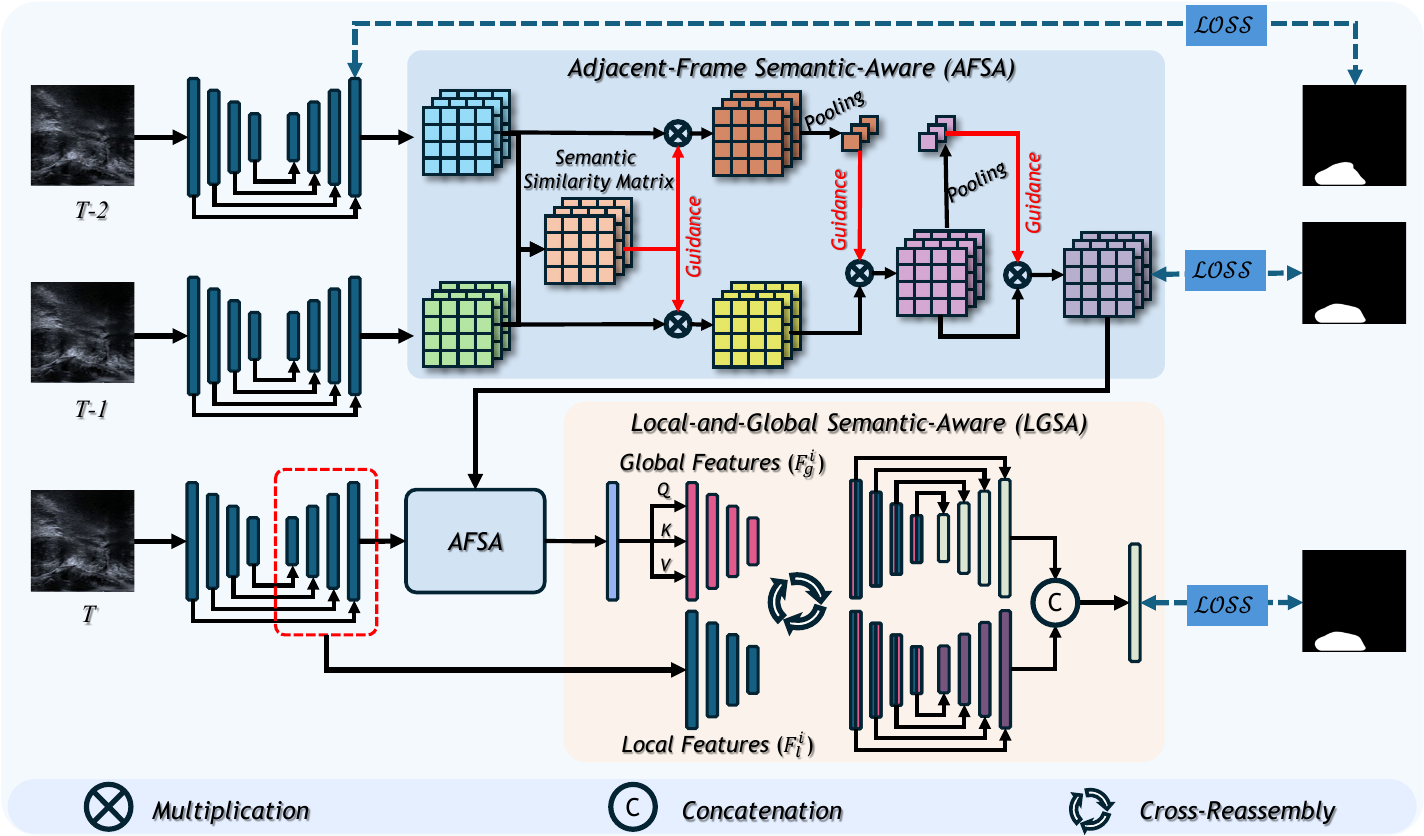}
\caption{Overview of the proposed Dual Semantic-Aware Network (DSANet). The Adjacent-Frame Semantic-Aware (AFSA) module captures channel-level correspondences between consecutive frames, while the Local-and-Global Semantic-Aware (LGSA) module fuses these cues to yield noise-robust segmentation maps. Segmentation losses are applied to the target frame and its two preceding frames, providing auxiliary temporal supervision.} \label{framework}
\end{figure*}

\section{Related Works}

Early advances in ultrasound segmentation predominantly centered on static image analysis. 
Jeremy \textit{et al.}\cite{webb2020automatic} introduced a spatio-temporal recurrent network that leverages temporal sequence information for automated segmentation of the thyroid gland in ultrasound cine clips. Gong \textit{et al.}
\cite{gong2023thyroid} proposed a multitask learning framework capable of concurrently segmenting both the thyroid gland and the nodule regions.

Despite these advances in image-based medical segmentation, such approaches generally lack the capacity to capture the dynamic evolution of lesions and organs—an aspect fundamental to informed clinical decision-making. In response, a growing body of research has shifted toward medical video segmentation, resulting in the emergence of more sophisticated techniques. For instance, Lin \textit{et al.}\cite{lin2023shifting} integrated local features with frequency-domain representations, demonstrating substantial improvements in breast lesion segmentation accuracy for ultrasound videos. Li \textit{et al.}\cite{li2022rethinking} developed a memory bank mechanism that dynamically captures and maintains both temporal and spatial correlations across video frames
Jiang \textit{et al.}\cite{jiang2024towards} analyzed semantic features in ultrasound videos by constructing an adaptive contextual sparse transformer, thereby facilitating effective modeling of complex spatiotemporal dependencies. Deng \textit{et al.}\cite{deng2024memsam} proposed a prompting strategy for the Segment Anything Model (SAM)\cite{kirillov2023segment}, integrating both spatial and temporal cues to enhance representation consistency and segmentation accuracy in echocardiography video analysis. 

While these methods have achieved promising results, they generally overlook the pervasive noise characteristics inherent to ultrasound video data. Distinct from these existing approaches, our method constructs fine-grained feature representations at the channel level, thereby enabling more effective suppression of intrinsic noise and leading to improved segmentation performance.

\section{Methods}
\label{Methods}
\subsection{Overview}

Fig. \ref{framework} provides a detailed illustration of our proposed DSANet framework. It processes a target frame along with its two preceding frames to explicitly leverage short-term temporal context. The backbone network extracts multi-scale features from each frame, which are then integrated using skip connections to preserve spatial detail. 

To address the challenges posed by speckle noise and temporal inconsistency in ultrasound data, we introduce two key components: the Adjacent-Frame Semantic-Aware (AFSA) module, which captures channel-level semantic correlations across frames, and the Local-and-Global Semantic-Aware (LGSA) module, which fuses unconditional local and conditional global features for noise-suppressed prediction. 
During training, segmentation losses are imposed not only on the target frame but also on the two preceding frames, providing auxiliary temporal supervision that enhances both spatial accuracy and temporal consistency. This dual-stage design enables DSANet to suppress speckle noise effectively, while achieving robust  segmentation and high computational efficiency.

\subsection{Adjacent-Frame Semantic-Aware}

\begin{figure*}[t]
\includegraphics[width=\linewidth]{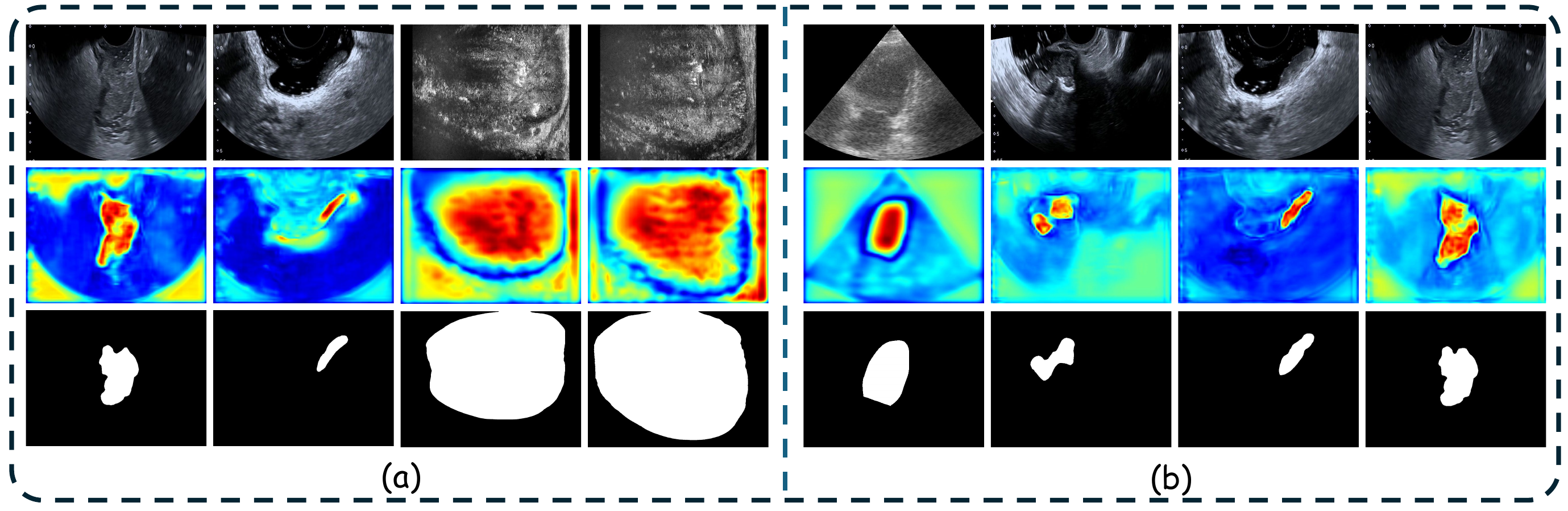}
\caption{(a) Heatmap visualization of the results generated by the AFSA module. (b) Heatmap visualization of the results generated by the LGSA module. From top to bottom: frames, heatmaps, and ground truth.} \label{Heatmap}
\end{figure*}

\subsubsection{Motivation}

In ultrasound video analysis,  speckle noise and subtle inter-frame appearance variations often compromise temporal feature stability. Although conventional pixel-level alignment methods, such as attention-based mechanisms, propagate features across frames, they are sensitive to pixel-wise intensity fluctuations and computationally expensive. 

To overcome these limitations, we propose the AFSA module, which performs lightweight temporal modeling by capturing semantic consistency at the channel level. This design is motivated by the fact that convolutional feature channels typically encode high-level semantic attributes, such as organ boundaries and the surrounding tissue context. While speckle noise manifests as local, irregular disturbances, its influence is inherently attenuated through spatial aggregation in convolution operations, and the aggregated semantic meaning within each channel remains relatively stable across adjacent frames. Leveraging this property, channel-wise correlation serves as an inherently noise-suppressed and efficient descriptor for modeling temporal consistency in ultrasound video sequences.

\subsubsection{Details}
As illustrated in Fig. \ref{framework}, AFSA takes feature maps from two consecutive frames, $F_t$ and $F_{t-1}$, as input. These are first reshaped from $\mathbb{R}^{C \times H \times W}$ to $\mathbb{R}^{C \times HW}$, where $C$ is the number of channels and $HW$ denotes the flattened spatial resolution. This reshaping operation enables a channel-wise perspective of spatial activations, allowing direct comparison of semantic responses across frames while intentionally abstracting away localized pixel fluctuations.

To model temporal correlations while suppressing random noise, a semantic similarity matrix $S \in \mathbb{R}^{C \times HW}$ is computed using cosine similarity between corresponding positions in the same channel across the two frames:
\begin{equation}
S_{i,j} = \frac{F_{t-1}[i,j] \cdot F_t[i,j]}{|F_{t-1}[i,j]||F_t[i,j]|}, \quad S \in \mathbb{R}^{C \times HW},
\end{equation}
where $F_{t-1}[i,j]$ and $F_t[i,j]$ represent the activations at spatial location $j$ in channel $i$ of each frame. This design leverages the inherent stability that semantic features encoded in the same channel across adjacent frames remain relatively stable, even in the presence of local intensity noise.

The semantic similarity matrix $S$ is then employed to guide the refinement of both frames. Specifically, $S$ emphasizes temporally consistent features by weighting $F_t$ and $F_{t-1}$ through element-wise multiplication:
\begin{equation}
F_t^{'} = S \cdot F_t, \quad
F_{t-1}^{'} = S \cdot F_{t-1},
\end{equation}
where $F_t^{'}$ and $F_{t-1}^{'}$ represent the refined features guided by the semantic similarity matrix. This operation enhances temporally reliable activations while attenuating unstable ones. By focusing on channel consistency, AFSA effectively preserves semantic continuity and suppresses local random noise.

To further integrate temporal context, AFSA applies global average pooling to the refined features. The pooled output of $F_{t-1}^{'}$ provides cross-frame temporal guidance, while that of $F_t^{'}$ serves as a self-context term. These are fused with $F_t^{'}$ through element-wise multiplication:
\begin{equation}
F_t^{final} = \operatorname{Pooling}(F_{t-1}^{'}) \cdot F_t^{'} \cdot \operatorname{Pooling}(F_t^{'}),
\end{equation}
yielding the final temporally enhanced representation. This design ensures that $F_t^{final}$ incorporates stable semantic cues from both itself and its preceding frame, achieving temporal coherence and improved noise robustness.

\subsubsection{Interpretability Analysis}

Fig. \ref{Heatmap} (a) presents heatmap visualizations of the results produced by the AFSA module. The heatmaps consistently and accurately highlight lesion regions and key anatomical structures, while effectively suppressing irrelevant background information. Although minor local speckles are present, no large-scale noise artifacts are observed. This clearly demonstrates the robust noise suppression capability of the proposed AFSA module under the challenging ultrasound video conditions.

\subsection{Local-and-Global Semantic-Aware}

\subsubsection{Motivation}

While temporal modeling enhances inter-frame feature consistency, comprehensive spatial semantic representations are equally crucial for effective ultrasound video segmentation. Without them, edge artifacts with blurry, jagged boundaries, such as those shown in Fig. \ref{Heatmap} (a), may arise. Given the presence of fine-grained anatomical details and broader contextual patterns, integrating local and global features efficiently is key to improving noise suppression and boundary localization. Inspired by prior work \cite{prajapati2021channel, liu2023tcgnet}, which demonstrated that channel-level feature construction is robust against noise and interference, we introduce the Local-and-Global Semantic-Aware (LGSA) module. This module fuses unconditional local features and conditional global features through a channel splitting and reassembling strategy, achieving multi-scale semantic enhancement and refined segmentation predictions.

\subsubsection{Details} As shown in Fig. \ref{framework}, the LGSA module comprises three key phases: global feature enhancement, channel reassembly fusion, and segmentation prediction.

{\bf Global features enhancement.} In this phase, the temporal conditional feature map $f_a = F_t^{final} \in \mathbb{R}^{C \times H \times W}$ obtained from the AFSA module is refined to emphasize semantically relevant information while suppressing residual noise. Specifically, a stacked self-attention mechanism is applied to capture long-range spatial dependencies within the target frame:
\begin{equation}
f_a^{SA} = \operatorname{softmax}\left(\frac{q(f_a),k(f_a)^T}{\sqrt{d}}\right)v(f_a),
\end{equation}
where $q(\cdot)$, $k(\cdot)$, and $v(\cdot)$ are linear projection functions, $\operatorname{softmax}(\cdot)$ denotes the softmax activation function and $d$ is the dimension of $f_a$. This process enables the network to adaptively assign higher weights to anatomically significant regions while down-weighting noisy or irrelevant areas, thus further enhancing the discriminative power of temporal conditional features. Then, the refined feature map $F_a^{SA}$ is passed through three convolutional layers to extract multi-scale fine-grained global features, denoted as $F_g^i$ for $(i=1, 2, 3, 4)$, with $F_g^1=F_a^{SA}$.

{\bf Channel reassembly fusion.}  Unconditional local features $F_l^i$ $ (i=1,2,3,4)$ extracted from earlier UNet-stages preserve frame-specific anatomical and holistic spatial information. These local features complement the dynamic, temporally enhanced global features $F_g^i$ by contributing spatial cues that may be attenuated during temporal modeling. To integrate these two types of information effectively, we perform a channel reassembly fusion operation. Specifically, for each scale $i$, the channels of $F_g^i$ and $F_l^i$ are split into two halves and reassembled through a cross-channel concatenation strategy:
\begin{equation}
\begin{split}
    F_{fus1}^i &= \operatorname{concat}(F_g^i[:C/2], F_l^i[C/2:]), \\
    F_{fus2}^i &= \operatorname{concat}(F_g^i[C/2:], F_l^i[:C/2]),
\end{split}
\end{equation}
where $C$ denotes the total number of channels per feature scale. This dual-path fusion strategy fosters mutual semantic awareness between unconditional local and conditional global features, encouraging the network to adaptively capture subtle anatomical variations while mitigating the influence of residual noise. By reassembling feature channels in complementary groupings, this operation enriches the feature diversity and ensures a balanced integration of both spatial and temporal cues at multiple scales.

{\bf Segmentation prediction.} In the final phase, the fused feature maps $F_{fus1}^i$ and $F_{fus2}^i$ are progressively upsampled to the original input resolution. The upsampled features are concatenated along the channel dimension, combining information from different fusion paths. Subsequently, a $1 \times 1$ convolutional layer is applied to generate the final segmentation prediction.

\subsubsection{Interpretability Analysis}

Fig. \ref{Heatmap} (b) shows heatmap visualizations of LGSA module. Compared to AFSA, LGSA outputs more precise attention on key anatomical regions corresponding to organs or lesions, with contours that are more complete and free from jagged edges, artifacts, or discontinuities. This indicates improved spatial awareness and semantic clarity resulting from the combined local-global fusion.

\subsection{Loss Function}

We employ a multi-term loss combining Dice, weighted binary cross-entropy (wBCE)\cite{wei2020f3net}, and weighted IoU (wIoU) \cite{wei2020f3net} to supervise our network. Specifically, we apply the loss not only to the target frame but also to its two preceding frames as auxiliary supervision. This design ensures that the network captures accurate spatial localization information, which is crucial for reliable temporal feature fusion. The total loss is defined as follows:

\begin{equation}
\begin{gathered}
    \mathcal{L}_{tar} = \mathcal{L}_{aux1} = \mathcal{L}_{aux2}
    = \mathcal{L}_{dice} + \mathcal{L}_{wbce} + \mathcal{L}_{iou}, \\
    \mathcal{L}_{total} = \mathcal{L}_{tar} + \mathcal{L}_{aux1} + \mathcal{L}_{aux2}.
\end{gathered}
\end{equation}

\section{Experiments}
\label{Experiments}

\subsection{Datasets and Evaluation Metrics}


We adopt four ultrasound video segmentation datasets to train
and evaluate our framework, including ERUS10K\cite{jiang2024towards}, Micro-Ultrasound Prostate Segmentation Dataset (Prostate)\cite{jiang2024microsegnet}, CAMUS\cite{leclerc2019deep} and VTUS\cite{10973086}. ERUS10K comprises 77 endorectal ultrasound videos, encompassing a total of 10,000 annotated frames. Prostate contains precisely annotated prostate ultrasound videos from 75 male patients. CAMUS contains high-quality cardiac ultrasound image data from 500 patients. VTUS consists of 100 thyroid ultrasound video sequences, each corresponding to an individual patient.

To better evaluate the performance of our proposed framework quantitatively, we adopt three common metrics, including Mean Absolute Error (MAE), Intersection of Union (IoU) and Dice Coefficient (Dice). Moreover, to demonstrate the inference efficiency of the model, we utilize inference Frame Per Second (FPS) as the measurement indicator.






\begin{table*}[t]
    \centering
    \caption{Quantitative comparison of DSANet and SOTA methods on four ultrasound‑video segmentation datasets. 
    {\bf BOLD} indicates the best result.}
    \label{tabComparison}
    \setlength{\tabcolsep}{4pt}
    \renewcommand{\arraystretch}{1.2}
    \resizebox{\textwidth}{!}{ 
    \begin{tabular}{@{}llcccccccccccccccc@{}}
        \toprule
        \multicolumn{2}{c}{} & \multicolumn{4}{c}{\textbf{EKUS10K} \cite{jiang2024towards}} & 
        \multicolumn{4}{c}{\textbf{Prostate} \cite{jiang2024microsegnet}} & 
        \multicolumn{4}{c}{\textbf{CAMUS} \cite{leclerc2019deep}} & \multicolumn{4}{c}{\textbf{VTUS} \cite{10973086}} \\
        \cmidrule(lr){3-6}\cmidrule(lr){7-10}\cmidrule(lr){11-14}\cmidrule(lr){15-18}
        \multicolumn{2}{c}{} & MAE$\downarrow$ & IoU$\uparrow$ & Dice$\uparrow$ & FPS$\uparrow$ 
                            & MAE$\downarrow$ & IoU$\uparrow$ & Dice$\uparrow$ & FPS$\uparrow$ 
                            & MAE$\downarrow$ & IoU$\uparrow$ & Dice$\uparrow$ & FPS$\uparrow$ & MAE$\downarrow$ & IoU$\uparrow$ & Dice$\uparrow$ & FPS$\uparrow$ \\
        \midrule
        \multicolumn{18}{l}{\textit{\textbf{Image‑based methods}}} \\[2pt]
        & UNet~\cite{ronneberger2015u}          & 4.0 & 55.0 & 68.9 & 182.9 & 7.1 & 83.2 & 89.6 & 194.6 & 2.3 & 82.4 & 90.1 & 235.6 &3.3&61.7&72.5&195.3\\
        & PraNet~\cite{fan2020pranet}           & 3.5 & 58.4 & 71.2 &  42.3 & 4.7 & 87.1 & 91.9 &  28.9 & 1.3 & 86.7 & 92.8 &  31.5 &2.7&68.6&79.1& 27.5\\
        & PFSNet~\cite{Ma_Xia_Li_2021}          & 3.5 & 50.4 & 61.7 &  34.2 & 4.6 & 86.4 & 92.1 &  35.7 & 1.3 & 87.4 & 93.2 &  29.5 &2.8&65.1&75.0&36.7\\
        & LDNet~\cite{zhang2022lesion}          & 3.5 & 59.1 & 72.4 &  31.5 & 4.9 & 85.4 & 91.1 &  24.3 & 1.3 & 86.7 & 92.8 &  21.3 &3.1&70.6&80.2& 27.1\\
        & LSSNet~\cite{wang2024lssnet}          & 3.5 & 60.1 & 72.7 &  16.9 & 4.9 & 86.1 & 91.6 &  16.6 & 1.3 & 86.8 & 92.8 &  15.3 &2.8&70.1&80.2& 19.0\\
        \addlinespace[2pt]
        \midrule
        \multicolumn{18}{l}{\textit{\textbf{Video‑based methods}}} \\[2pt]
        & PNSNet~\cite{ji2021progressively}     & 3.7 & 58.9 & 72.5 &  28.2 & 5.2 & 84.2 & 90.1 &  26.0 & 1.5 & 85.6 & 92.1 &  30.7 &4.2&64.6&74.7&33.8\\
        & SLTNet~\cite{cheng2022implicit}       & 4.1 & 45.4 & 58.8 &   4.3 & 7.3 & 79.9 & 87.9 &   4.2 & 1.7 & 83.3 & 90.7 &   4.5 &6.8&60.2&72.1&4.7\\
        & FLANet~\cite{lin2023shifting}         & 3.8 & 58.0 & 71.5 &  30.9 & 5.6 & 86.0 & 90.9 &  33.5 & 1.7 & 84.5 & 91.5 &  28.7 &2.9&71.4&81.3& 33.0\\
        & SALI~\cite{hu2024sali}                & \textbf{3.1} & 61.9 & 75.0 &   6.0 & 4.7 & 86.6 & 92.1 &   7.9 & 1.4 & 86.4 & 92.6 &7.2 &2.9&69.6&79.7& 6.7\\
        & ASTR~\cite{jiang2024towards}          & 3.2 & 62.2 & 75.7 &  16.4 & 6.9 & 85.8 & 91.0 &  23.5 & 1.5 & 85.7 & 92.2 &  24.2 &2.9&70.6&81.0&28.2 \\
        & SAM2~\cite{ravi2024sam}   & 3.9& 60.2 & 72.2 & 7.8 & 8.9  &85.4  & 91.0 & 7.8 & 1.56  & 85.4 & 91.9 & 8.4 & 4.2  &70.2&80.4&8.0\\
        & MedSAM2~\cite{zhu2024medical}  & 3.8 & 62.6 & 73.8& 10.6 & 5.9  & 86.2 & 91.6 & 10.4 &  1.5 & 86.5 & 92.7 & 11.6 &  3.8 &71.9&81.8&10.7\\
        \rowcolor{gray!15}
        & \textbf{Ours}                         & \textbf{3.1} & \textbf{63.3} & \textbf{76.5} & \textbf{33.8}& \textbf{4.0} & \textbf{87.6} & \textbf{92.5} & \textbf{36.9}& \textbf{1.2} & \textbf{88.0} & \textbf{93.5} & \textbf{38.3} &\textbf{2.6}&\textbf{72.4}&\textbf{82.3}&\textbf{34.5}\\
        \bottomrule
    \end{tabular}}
\end{table*}

\begin{figure*}[!h]
\includegraphics[width=\textwidth]{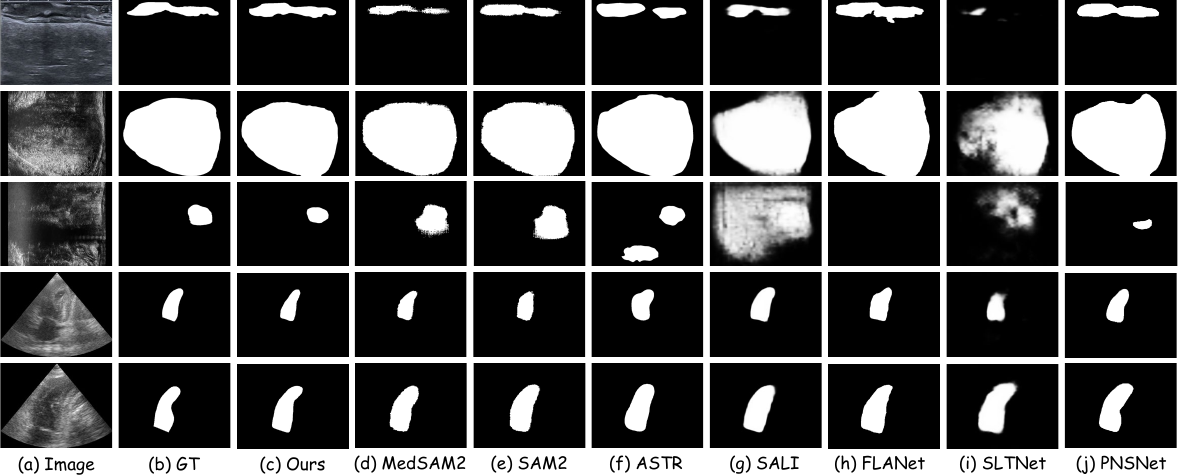}
\caption{Qualitative comparison of DSANet with video-based SOTA methods.} \label{figcomparison}
\end{figure*}

\subsection{Implementation Details}
The proposed framework is implemented using PyTorch and trained for 15 epochs with a batch size of 2 on a RTX 3090 GPU. We adopt Res2Net-50 \cite{gao2019res2net}, pretrained on ImageNet, as the backbone. During training and inference, a video clip of $T=3$ frames is selected. All input images are resized to 352 $\times$ 352 resolution and augmented with random flipping. Optimization is performed using Adam optimizer with an initial learning rate of 0.0001 and weight decay of 0.0005.

\begin{table*}[t]
    \centering
    \caption{Ablation studies on effectiveness and noise robustness of different components. {\bf BOLD} indicates the best result.}
    \label{tab:component}
    \setlength{\tabcolsep}{5pt}  
    \renewcommand{\arraystretch}{1.25}  
    \resizebox{\textwidth}{!}{
    \begin{tabular}{lcccccccccccc}
        \toprule
        & \multicolumn{3}{c}{\textbf{ERUS10K}}
        & \multicolumn{3}{c}{\textbf{Prostate}}
        & \multicolumn{3}{c}{\textbf{CAMUS}}
        & \multicolumn{3}{c}{\textbf{VTUS}} \\
        \cmidrule(lr){2-4}\cmidrule(lr){5-7}\cmidrule(lr){8-10}\cmidrule(lr){11-13}
        & MAE$\downarrow$ & IoU$\uparrow$ & Dice$\uparrow$
        & MAE$\downarrow$ & IoU$\uparrow$ & Dice$\uparrow$
        & MAE$\downarrow$ & IoU$\uparrow$ & Dice$\uparrow$
        & MAE$\downarrow$ & IoU$\uparrow$ & Dice$\uparrow$ \\
        \midrule
        \rowcolor{gray!15}
        \multicolumn{13}{l}{\textit{\textbf{Baseline}}} \\
        & 3.6 & 58.6 & 71.7 & 6.5 & 84.1 & 90.2 & 1.7 & 84.9 & 91.7 & 4.5 & 62.2 & 73.7\\
        $L=25$ & $4.2_{(\uparrow0.6)}$ & $54.7_{(\downarrow3.9)}$ & $68.6_{(\downarrow3.1)}$
               & $7.4_{(\uparrow0.9)}$ & $81.7_{(\downarrow2.4)}$ & $88.1_{(\downarrow2.1)}$
               & $2.1_{(\uparrow0.4)}$ & $81.3_{(\downarrow3.6)}$ & $89.5_{(\downarrow2.2)}$
               & $4.9_{(\uparrow0.4)}$ & $59.4_{(\downarrow2.8)}$ & $71.5_{(\downarrow2.2)}$\\
        $L=20$ & $4.6_{(\uparrow1.0)}$ & $51.9_{(\downarrow6.5)}$ & $66.4_{(\downarrow5.3)}$
               & $9.8_{(\uparrow3.3)}$ & $79.9_{(\downarrow4.2)}$ & $86.5_{(\downarrow3.7)}$
               & $2.4_{(\uparrow0.7)}$ & $80.1_{(\downarrow4.8)}$ & $88.7_{(\downarrow3.0)}$
               & $6.7_{(\uparrow2.2)}$ & $59.2_{(\downarrow3.0)}$ & $70.0_{(\downarrow3.7)}$\\
        \rowcolor{gray!15}
        \multicolumn{13}{l}{\textit{\textbf{Baseline + AFSA}}} \\
        & 3.7 & 60.1 & 73.8 & 5.2 & 85.9 & 90.9 & 1.6 & 85.3 & 92.0 & 3.5 & 67.9 & 78.5\\
        $L=25$ & $4.1_{(\uparrow0.4)}$ & $59.3_{(\downarrow0.8)}$ & $72.6_{(\downarrow1.2)}$
               & $6.3_{(\uparrow1.1)}$ & $85.2_{(\downarrow0.7)}$ & $90.3_{(\downarrow0.6)}$
               & $1.8_{(\uparrow0.2)}$ & $84.0_{(\downarrow1.3)}$ & $91.2_{(\downarrow1.2)}$
               & $3.8_{(\uparrow0.3)}$ & $65.9_{(\downarrow2.0)}$ & $77.1_{(\downarrow1.4)}$\\
        $L=20$ & $4.1_{(\uparrow0.4)}$ & $58.0_{(\downarrow2.1)}$ & $72.0_{(\downarrow1.8)}$
               & $6.3_{(\uparrow1.1)}$ & $85.0_{(\downarrow0.9)}$ & $90.2_{(\downarrow0.7)}$
               & $1.9_{(\uparrow0.3)}$ & $83.0_{(\downarrow2.3)}$ & $90.4_{(\downarrow1.4)}$
               & $4.1_{(\uparrow0.6)}$ & $65.9_{(\downarrow2.0)}$ & $76.8_{(\downarrow1.7)}$\\
        \rowcolor{gray!15}
        \multicolumn{13}{l}{\textit{\textbf{Baseline + LGSA}}} \\
        & 3.4 & 60.5 & 74.1 & 4.9 & 85.8 & 90.8 & 1.6 & 85.8 & 92.3 & 3.5 & 67.9 & 78.5\\
        $L=25$ & $3.8_{(\uparrow0.4)}$ & $59.5_{(\downarrow1.0)}$ & $72.4_{(\downarrow1.7)}$
               & $5.5_{(\uparrow0.6)}$ & $85.0_{(\downarrow0.8)}$ & $90.7_{(\downarrow0.1)}$
               & $1.7_{(\uparrow0.1)}$ & $83.6_{(\downarrow2.2)}$ & $91.0_{(\downarrow1.3)}$
               & $3.6_{(\uparrow0.1)}$ & $66.1_{(\downarrow1.8)}$ & $76.9_{(\downarrow1.6)}$\\
        $L=20$ & $3.8_{(\uparrow0.4)}$ & $58.9_{(\downarrow1.6)}$ & $72.1_{(\downarrow2.0)}$
               & $5.9_{(\uparrow1.0)}$ & $84.8_{(\downarrow1.0)}$ & $90.7_{(\downarrow0.1)}$
               & $1.8_{(\uparrow0.2)}$ & $83.6_{(\downarrow2.2)}$ & $90.9_{(\downarrow1.4)}$
               & $3.9_{(\uparrow0.4)}$ & $65.8_{(\downarrow2.1)}$ & $76.5_{(\downarrow2.0)}$\\
        \rowcolor{gray!15}
        \multicolumn{13}{l}{\textit{\textbf{Full model}}} \\
        & \textbf{3.2} & \textbf{63.2} & \textbf{76.4}
        & \textbf{4.5} & \textbf{87.6} & \textbf{92.5}
        & \textbf{1.2} & \textbf{88.0} & \textbf{93.5}
        & \textbf{2.6} & \textbf{72.4} & \textbf{82.3} \\
        \bottomrule
    \end{tabular}}
\end{table*}

\begin{figure}[t]
\includegraphics[width=\linewidth]{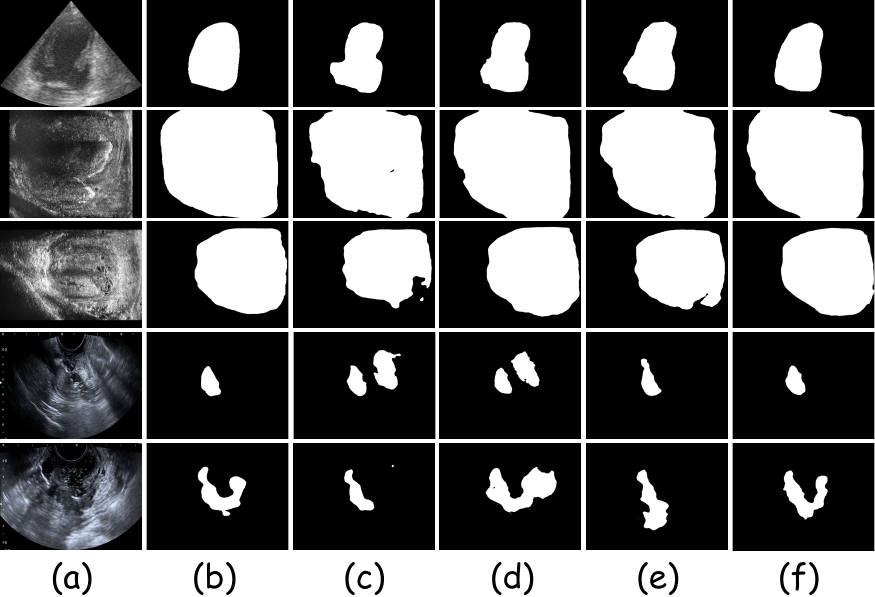}
\caption{Visual Comparison of component effectiveness. (a) Frame, (b) Ground Truth, (c) Baseline, (d) Baseline + AFSA, (e) Baseline + LGSA, (f) Full Model.} \label{comcom}
\end{figure}

\subsection{Comparison with State-of-the-Art Methods}

To better evaluate the performance of our framework, we compare it against 12 SOTA methods, including 5 image-based methods: UNet \cite{ronneberger2015u}, PraNet \cite{fan2020pranet}, PFSNet \cite{Ma_Xia_Li_2021}, LDNet \cite{zhang2022lesion} and LSSNet \cite{wang2024lssnet}, and 7 video-based methods: PNSNet \cite{ji2021progressively}, SLTNet \cite{cheng2022implicit}, FLANet \cite{lin2023shifting}, SALI \cite{hu2024sali}, ASTR \cite{jiang2024towards}, SAM2\cite{ravi2024sam}, and MedSAM2\cite{zhu2024medical}.

\subsubsection{Quantitative Comparisons}

Tab.~\ref{tabComparison} reports the quantitative results of different models in terms of MAE, IoU, Dice, and FPS. Obviously, our proposed method consistently achieves the best performance across all four datasets. Notably, it outperforms existing SOTA methods by nearly 1\% in both IoU and Dice scores. 

In terms of efficiency, our model surpasses the FPS of all video-based methods, with improvements of 5.6\%, 2.4\%, 7.6\% and 0.7\% on the ERUS10K, Prostate, CAMUS and VTUS datasets, respectively, and even exceeds several image-based approaches. These results demonstrate the high segmentation accuracy and real-time inference speed of our method.

\subsubsection{Qualitative Comparisons}

Fig. \ref{figcomparison} illustrates qualitative  comparisons between our approach and video-based SOTA methods. 
Our model demonstrates a clear advantage by  accurately localizing the targets and detecting their boundaries. In contrast, competing methods often fail to handle the impact of noise properly, resulting in boundary ambiguity (e.g. methods (d) \& (e)), segmentation failure (e.g. methods (f) \& (g)), and incomplete segmentation (e.g. methods (h), (i) \& (j)). 
These visual results confirm that our proposed dual semantic-aware design enhances noise robustness and spatial consistency, yielding more reliable segmentation outcomes under real-world ultrasound video conditions. 

\subsection{Ablation Studies}
\subsubsection{Effectiveness of Components}


Table \ref{tab:component} presents the quantitative comparison of various architectural components. The baseline model achieves a Mean Absolute Error (MAE) of 3.6, an Intersection over Union (IoU) of 58.6\%, and a Dice coefficient of 71.7\% on the EKUS10K dataset. Incorporating the AFSA module enhances the IoU to 60.1\% and the Dice score to 73.8\%, demonstrating its capacity to capture temporal dependencies and mitigate noise. The addition of the LGSA module, which integrates both temporal conditional global and unconditional local features, further improves performance, increasing the IoU to 60.5\% and Dice to 74.1\%. When both the AFSA and LGSA modules are combined, the model achieves the highest performance, with IoU and Dice reaching 63.2\% and 76.4\%, respectively, on EKUS10K. Similar improvements are observed on the Prostate, CAMUS, and VTUS datasets.




As shown in Fig. \ref{comcom}, the baseline model frequently produces incomplete segmentations (e.g., rows 1 and 5), ambiguous boundaries (e.g., rows 2 and 3), and incorrect predictions (e.g., row 4). These issues are primarily caused by speckle noise inherent in ultrasound videos, which corrupts latent feature representations and impairs the model's ability to localize target regions. 
In contrast, although the combinations of baseline with AFSA or LGSA still exhibit occasional segmentation inaccuracies, they significantly mitigate these issues by producing sharper boundaries and more complete segmentation maps. This improvement can be attributed to our method's ability to suppress noise interference at the channel level, thereby enhancing the reliability of the learned feature representations. When integrated, the joint effect of AFSA and LGSA enables the model to generate segmentation maps close to the ground truth. In summary, both AFSA and LGSA significantly contribute to speckle noise suppression and improved segmentation in ultrasound video. 

\subsubsection{Performance of Noise Suppression}

To evaluate the noise robustness, we simulate varying speckle noise intensities using a multiplicative Gamma noise model, a standard method for mimicking ultrasound speckle artifacts. In our setting, we consider three conditions: 
(1) original ultrasound videos, 
(2) mildly enhanced speckle noise with a reduced look number $L=25$, and 
(3) heavily enhanced speckle noise with $L=20$. 

Quantitative results under increasing noise conditions are presented in Tab. \ref{tab:component}. Without AFSA and LGSA, the baseline model's segmentation accuracy sharply declines under noise. Specifically, when the look number is reduced to $L=25$, performance deteriorates significantly, with Dice scores on all datasets dropping by about 3.0 points. As the look number decreases further to $L=20$, performance worsens, highlighting the model's vulnerability to increased speckle noise.

In contrast, integrating AFSA and LGSA significantly enhances noise robustness. Performance degradation across all datasets is notably reduced, with these three metrics remaining more stable under both noise levels ($L=25$ and $L=20$). These results confirm that the proposed modules strengthen the model's resilience to speckle noise, effectively maintaining segmentation accuracy under challenging imaging conditions.

\subsubsection{Effectiveness of Local-and-Global Strategy}

\begin{figure}[t]
\includegraphics[width=\linewidth]{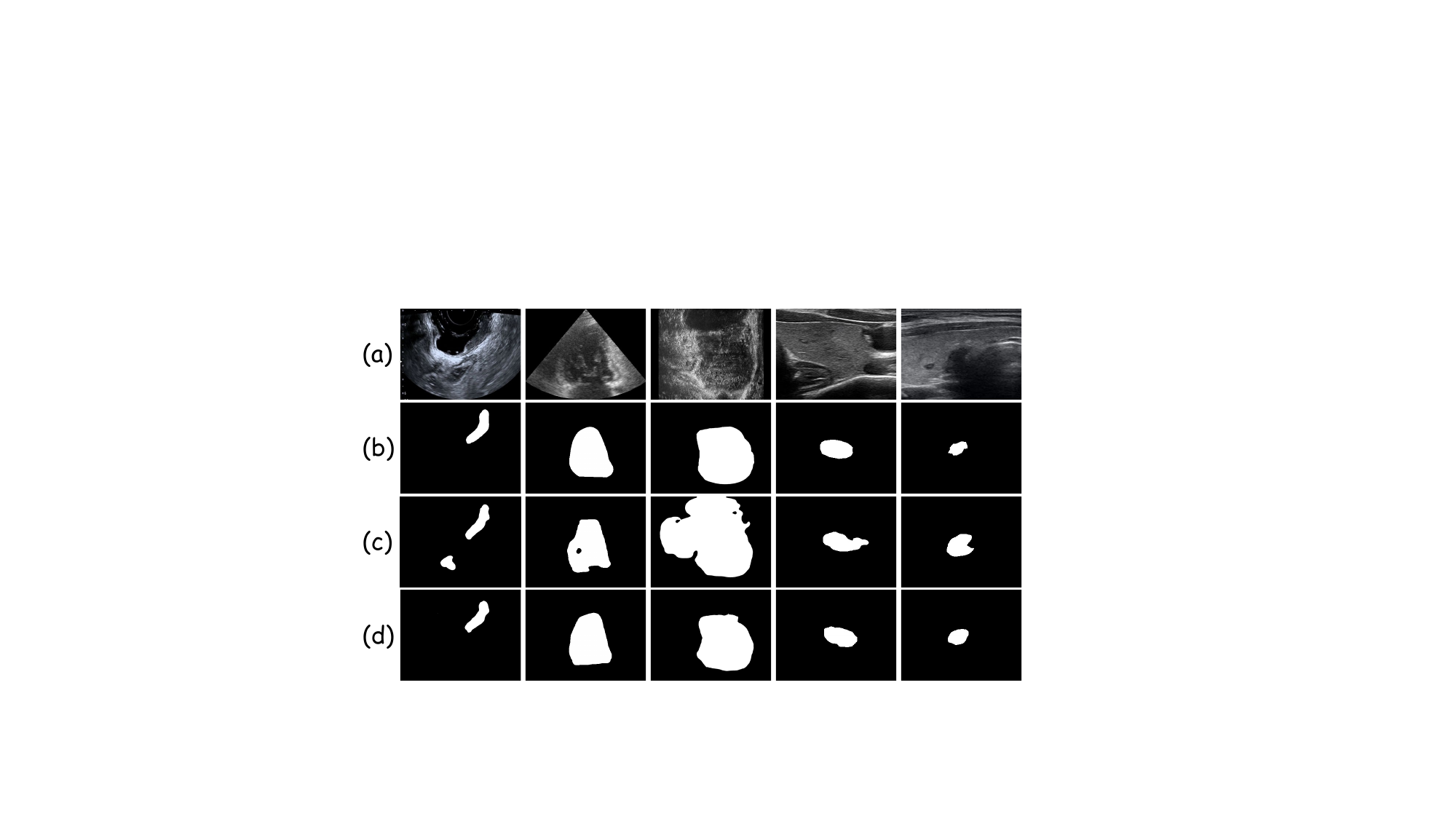}
\caption{Visual Comparison between direct global and local-and-global strategies. (a) Frame, (b) Ground Truth, (c) Direct global, (d) Local-and-Global.} \label{ltg}
\end{figure}

The proposed local-and-global strategy is designed to mitigate noise interference during temporal fusion by reintroducing complementary local spatial cues. In contrast, relying solely on a global strategy tends to disrupt latent feature representations, resulting in inaccurate and unstable segmentation outcomes.

To validate the effectiveness of this strategy, we perform ablation studies comparing a conventional direct global fusion scheme with our local-and-global fusion framework. As illustrated in Fig. \ref{ltg}, direct global fusion tends to produce segmentation results with redundant or over-segmented regions, as it lacks the complementary spatial detail provided by local features. 
In contrast, our local-and-global strategy yields precise segmentation boundaries that closely match the ground truth. These results demonstrate that integrating frame-specific spatial cues into the fusion process not only improves segmentation precision but also reduces noise-induced artifacts.



\section{Conclusion}
\label{Conclusion}

In this paper, we introduce DSANet, a novel dual semantic-aware network for noise-robust ultrasound video segmentation. Our approach integrates two key modules, AFSA and LGSA, both designed to mitigate inherent ultrasound noise and improve segmentation accuracy. We evaluate our model against 5 image-based and 7 video-based SOTA segmentation methods on four benchmark datasets. Experimental results demonstrate the superior performance and robustness of our model under challenging conditions. This work advances ultrasound video lesion segmentation and supports automated diagnostics, enabling more reliable and efficient clinical applications.






\end{document}